\documentclass[pdflatex,sn-mathphys-num]{sn-jnl}

\usepackage{graphicx}%
\usepackage{multirow}%
\usepackage{amsmath,amssymb,amsfonts}%
\usepackage{amsthm}%
\usepackage{mathrsfs}%
\usepackage[title]{appendix}%
\usepackage{xcolor}%
\usepackage{textcomp}%
\usepackage{manyfoot}%
\usepackage{booktabs}%
\usepackage{algorithm}%
\usepackage{algorithmicx}%
\usepackage{algpseudocode}%
\usepackage{listings}%

\theoremstyle{thmstyleone}%
\theoremstyle{thmstyletwo}%

\theoremstyle{thmstylethree}%

\begin{document}

\title[Article Title]{Multi-Stage Generative Upscaler: Reconstructing Football Broadcast Images via Diffusion Models}


\author*[1]{\fnm{Luca} \sur{Martini}}\email{luca.martini@edu.unige.it}

\author[2]{\fnm{Daniele} \sur{Zolezzi}}\email{daniele.zolezzi@edu.unige.it}
\equalcont{These authors contributed equally to this work.}

\author[2]{\fnm{Saverio} \sur{Iacono}}\email{saverio.iacono@unige.it}
\equalcont{These authors contributed equally to this work.}

\author[2]{\fnm{Gianni Viardo} \sur{Vercelli}}\email{gianni.vercelli@unige.it}
\equalcont{These authors contributed equally to this work.}

\affil*[1]{\orgdiv{Department of Languages and Modern Culture}, \orgname{University of Genova}, \orgaddress{\city{Genova}, \postcode{16124}, \country{Italy}}}

\affil[2]{\orgdiv{Department of Computer Science and Technology, Bioengineering, Robotics, and Systems Engineering}, \orgname{University of Genova}, \orgaddress{\city{Genova}, \postcode{16145}, \country{Italy}}}

\abstract{The reconstruction of low-resolution football broadcast images presents a significant challenge in sports broadcasting, where detailed visuals are essential for analysis and audience engagement. This study introduces a multi-stage generative upscaling framework leveraging Diffusion Models to enhance degraded images, transforming inputs as small as $64 \times 64$ pixels into high-fidelity $1024 \times 1024$ outputs. By integrating an image-to-image pipeline, ControlNet conditioning, and LoRA fine-tuning, our approach surpasses traditional upscaling methods in restoring intricate textures, sharp edges, and domain-specific elements such as player details and jersey logos. The custom LoRA is trained on a custom football dataset, ensuring adaptability to sports broadcast needs. Experimental results demonstrate substantial improvements over conventional models, with ControlNet refining fine details and LoRA enhancing task-specific elements. These findings highlight the potential of diffusion-based image reconstruction in sports media, paving the way for future applications in automated video enhancement and real-time sports analytics.}




\maketitle

\section{Introduction}\label{sec1}

Generative Artificial Intelligence (genAI) represents a groundbreaking approach to creativity and automation, empowering machines to produce novel and highly realistic data, including images, text, and music. Among the diverse generative models, Diffusion Models have emerged as a powerful technique for high-quality image synthesis. Rooted in the principles of probabilistic modeling, Diffusion Models iteratively refine noise into detailed and coherent representations, achieving remarkable performance in domains like image generation, image inpainting and style transfer.

Diffusion Models have gained traction due to their versatility and robustness, allowing them to excel in challenging tasks where conventional generative approaches, such as Generative Adversarial Networks (GANs), often struggle. These models leverage a forward-backward diffusion process, where images are progressively noised during the forward phase and restored to their original form during the reverse phase. This framework not only provides a stable training environment but also generates results with exceptional fidelity and diversity.

In this study, we focus on leveraging Diffusion Models to reconstruct and upscale low-resolution images, transforming inputs as small as $64 \times 64$ pixels into high-resolution outputs of $1024 \times 1024$ pixels or greater. This innovative approach goes beyond merely augmenting image resolution; it also reconstructs critical details in areas of the image that are heavily blurred, noisy, or otherwise degraded. The process enables the restoration of fine-grained textures and intricate visual elements, which are often missing or indistinct in the original low-resolution images.

The work presented in this study has been conducted in collaboration with a broadcast company specializing in sports events, particularly football, including major tournaments such as the UEFA Champions League. A significant challenge in this domain arises from low-resolution images generated as cutouts from live camera broadcasts. These cutouts are often used in post-match analysis, where detailed visual content is essential for reviewing critical moments, analyzing player movements, and creating engaging commentary. However, the quality of these images is frequently compromised due to cropping, compression, and real-time processing constraints.

Our approach addresses this challenge by reconstructing and enhancing these low-resolution cutouts, restoring general image structure, specific details and improving overall clarity. This enables broadcasters to deliver high-quality visuals that meet the exacting standards required for detailed analysis and audience satisfaction, while also preserving the authenticity and integrity of the original footage.

To achieve this reconstruction, we leverage the Diffusers library \cite{huggingface_diffusers} and Flux.1-Dev model \cite{black_forest_labs_website}, employing a systematic multi-stage process to maximize reconstruction accuracy and detail recovery. In the first stage, an image-to-image pipeline is used to reconstruct broader and general details of the low-resolution inputs, providing a foundational enhancement. Following this, we employ a pre-trained ControlNet model \cite{Jasperai/flux.1-dev-controlnet-upscaler} to augment finer details and improve the fidelity of the reconstructed images. Finally, to further specialize the reconstruction capabilities of ControlNet in the context of football broadcasts, we train a LoRA (Low-Rank Adaptation) \cite{LoRA_hu2022_LLMs} model on a handcrafted football dataset. This targeted fine-tuning enables ControlNet model to better capture and reconstruct domain-specific elements, such as player details, jerseys, and field textures, ensuring the enhanced images meet the high-quality standards demanded in sports broadcasting workflows.

The paper presents the work with: 
\begin{itemize}
    \item Section 2 (Related Works), the study reviews prior advancements in image reconstruction, including GAN-based techniques like SRGAN and ESRGAN, and contrasts them with the emerging superiority of Diffusion Models.
    \item Section 3 (Methodology) details the reconstruction framework in the following manner:
    \begin{itemize}
        \item Section 3.1 (First Stage of Reconstruction), where input images are standardized and initial upscaling is performed using an image-to-image pipeline.
        \item Section 3.2 (Second Stage of Reconstruction) introduces ControlNet to refine finer details like player numbers and jersey textures.
        \item Section 3.3 (Fine-Tuning) describes the custom LoRA model trained on a football-specific dataset to further enhance domain-specific elements.
    \end{itemize}
    \item Section 4 (Results) evaluates the effectiveness of the proposed approach, demonstrating improved clarity and structural fidelity, particularly with the integration of LoRA.
    \item Section 5 (Conclusions) discusses future improvements, including dataset expansion, Graphical User Interface development and additional fine-tuning strategies for further refinement. 
\end{itemize}

\section{Related Works}\label{sec2}

The field of image reconstruction and enhancement has seen significant advancements in recent years, driven by the development of powerful generative models. Among these, Diffusion Models have emerged as a leading technique, offering superior performance in generating high-quality images from noisy or degraded inputs. Building on the foundational work of ``Denoising Diffusion Probabilistic Models'' (DDPM) \cite{ho_2020_denoising}, Diffusion Models have demonstrated remarkable versatility across different domain such as inpainting and image synthesis. Their probabilistic framework, which iteratively refines data through a noise-to-data reversal process, has set a new benchmark for generative modeling.

Among the most notable advancements in image reconstruction and enhancement, Generative Adversarial Networks (GANs) have revolutionized the field with their ability to generate perceptually convincing images. The introduction of the Super-Resolution Generative Adversarial Network (SRGAN) \cite{SRGAN2017} marked a turning point in single-image super-resolution tasks. SRGAN employed a novel perceptual loss function that combined content loss, based on high-level feature maps from a pre-trained VGG (Visual Geometry Group) network \cite{VGGNet2014}, and adversarial loss to enhance texture details and generate images with realistic characteristics. This architecture demonstrated a significant improvement over traditional methods by focusing not only on pixel-wise accuracy but also on generating visually appealing results, even when the input images were highly degraded.

Building on SRGAN, Enhanced Super-Resolution GAN (ESRGAN) introduced by \cite{ESRGAN2019} further advanced the state of the art by addressing some of SRGAN's inherent shortcomings. ESRGAN replaced the vanilla residual blocks in SRGAN with Residual-in-Residual Dense Blocks (RRDB), which improved gradient flow and enabled deeper network architectures. Furthermore, ESRGAN incorporated a relativistic adversarial loss, which improved the discriminator's ability to distinguish between real and fake samples, leading to more stable training dynamics. These innovations resulted in sharper textures, better restoration of fine details, and a noticeable improvement in the perceptual quality of super-resolved images.

Despite the significant improvements made by SRGAN and ESRGAN, their performance often faltered when applied to real-world scenarios involving diverse degradations not seen during training. To address this gap, Real-ESRGAN \cite{RealESRGAN2021} was introduced, extending the capabilities of GAN-based approaches to handle real-world blind super-resolution tasks. Real-ESRGAN enhanced ESRGAN's architecture by introducing a pure synthetic data generation pipeline for training, which simulated a wide variety of real-world degradations, such as noise, blur, compression artifacts, and downsampling effects. This approach significantly increased the robustness of the model to unseen degradations. Real-ESRGAN also incorporated a second-order degradation model to better approximate real-world data characteristics, along with adaptive weight strategies to balance training between perceptual quality and pixel fidelity. These modifications allowed Real-ESRGAN to produce high-quality outputs that maintained both visual realism and structural accuracy, even under challenging conditions. By emphasizing the diversity and realism of its training data, Real-ESRGAN outperformed its predecessors in tasks where input images suffered from significant and unpredictable degradations.

Despite the improvements brought by ESRGAN and Real-ESRGAN, the need for a more robust approach capable of handling extreme upscaling and severe image degradations prompted the exploration of alternative frameworks. For our scope of work, these limitations of GAN-based approaches prompted the exploration of Diffusion Models, which offer stable training dynamics and the ability to generate high-fidelity details across extreme upscaling tasks. Diffusion Models' iterative denoising process is particularly well-suited for reconstructing not only resolution but also missing or blurred elements, providing a more robust solution for reconstructing low-resolution, degraded camera cutouts used in post-match analyses of sports events.
Notably, the use of Diffusion Models extends beyond generative tasks to support practical applications in image restoration and enhancement. A recent study \cite{SROverview2023} highlighted the use of Diffusion Models in addressing extreme challenges in image reconstruction by leveraging their iterative refinement mechanism. This method integrates degradation-specific conditioning to address noise, blur, and downscaling artifacts effectively. The study underscores the diffusion framework's adaptability in learning fine-grained details, surpassing traditional and GAN-based approaches in reconstructing visually coherent high-resolution outputs.

Advancements in the capabilities of Diffusion Models have been significantly enhanced by frameworks such as ControlNet, which refine the generative process through the integration of conditional control. The groundbreaking work, \textit{Adding Conditional Control to Text-to-Image Diffusion Models} \cite{ControlNet_Zhang_2023_ICCV} propose a novel method to incorporate external conditioning signals, such as edge maps, depth information, or pose guidance, into the generation process. This innovation not only improves the flexibility of Diffusion Models but does so without compromising the pre-trained generative capacity of the underlying model.
ControlNet introduces a dedicated control branch tailored to process supplementary conditioning inputs while maintaining the original weights of the pre-trained diffusion model. By freezing the core model parameters and focusing training solely on the control branch, this architecture achieves an optimal trade-off between model fine-tuning and stability. This design ensures that the diffusion model preserves its ability to generate high-quality images while enabling precise control over specific aspects of image generation.
In comparison to conventional approaches, ControlNet circumvents the need for extensive parameter adjustments, which are often necessary in GAN-based or fully fine-tuned Diffusion Models. Its lightweight and modular framework makes it particularly effective for iterative enhancements in specialized applications, such as real-time sports analysis or domain-specific image restoration.

Building on the advancements introduced by ControlNet, which enable precise conditional control in Diffusion Models, the need for efficient adaptation of such models to diverse domains becomes increasingly evident. While ControlNet provides a framework for integrating external conditioning data, the process of fine-tuning Diffusion Models for specific applications still poses significant computational challenges. This highlights the need for approaches that enable lightweight, task-specific modifications without retraining the entire model, a challenge effectively addressed by techniques like Low-Rank Adaptation (LoRA) \cite{LoRA_hu2022_LLMs}.
Originally introduced as a technique for efficiently fine-tuning Large Language Models, LoRA has found remarkable relevance in the domain of Diffusion Models \cite{Lora_Diffusers2023}. LoRA addresses one of the core challenges in training and adapting large models: the computational cost and memory requirements of updating a vast number of parameters. By introducing a low-rank decomposition framework, LoRA enables the adjustment of model weights in a lightweight and scalable manner, making it an attractive alternative for adapting Diffusion Models to specific tasks or datasets. Furthermore, LoRA’s approach of isolating task-specific modifications ensures that the underlying generative capabilities of the diffusion model remain intact, reducing the risk of catastrophic forgetting. This makes LoRA particularly suited for scenarios where Diffusion Models need to handle diverse tasks or datasets with minimal retraining. Recent studies, further highlight the superiority of LoRA over other fine-tuning techniques like Textual Inversion or Dreambooth \cite{martini_2024}, emphasizing its efficiency and versatility across a wide range of applications.
Notably, LoRA has also demonstrated superior computational efficiency, often enabling fine-tuning to be performed even on consumer-grade hardware, making it accessible for a wider range of applications and researchers \cite{martini_2024}.

Recent developments in diffusion model architecture have marked a shift from the traditional reliance on U-Net backbones to Transformer-based designs called Diffusion Transformers (DiT) \cite{DiT_peebles2023}. While U-Nets have been the backbone of choice due to their convolutional efficiency and hierarchical feature extraction, they often struggle with capturing long-range dependencies and handling intricate textures. Transformer-based architectures address these limitations through self-attention mechanisms, which enable the modeling of global dependencies and maintain contextual coherence throughout the generation process. These characteristics make Transformers particularly suitable for complex tasks requiring the reconstruction of fine details or the synthesis of high-resolution outputs.

Simultaneously, the Rectified Flow \cite{RectifiedFlow_yan2023} approach has emerged as a significant innovation, enhancing the efficiency and stability of Diffusion Models. Rectified Flow optimizes the noise-to-data reversal process by aligning the transport paths more closely with the shortest trajectories between data distributions. This alignment reduces gradient instability, which is a common issue in traditional models, and ensures smoother training dynamics. Furthermore, Rectified Flow adapts dynamically to the complexity of input data, enabling robust handling of diverse degradations, such as noise, blur, and compression artifacts. These properties allow Rectified Flow to generate high-quality images with fewer inference steps, making it computationally efficient while maintaining output fidelity.

In conclusion, for the scope of our work, we leverage state-of-the-art (SOTA) advancements in Diffusion Models to tackle the challenges of image reconstruction and enhancement. Our approach incorporates the latest models featuring DiT and Rectified Flow architectures to ensure efficiency and stability in the generative process. Additionally, we utilize ControlNet for precise conditional control and LoRA for lightweight and efficient fine-tuning, enabling task-specific adaptations with minimal computational overhead. This integration of cutting-edge techniques positions our methodology to address diverse image degradation scenarios effectively, setting a robust foundation for high-fidelity reconstruction tasks.

\section{Methodology}\label{sec3}

This study focuses on developing a robust framework to enhance the resolution of low-quality images and reconstruct severely degraded regions, aiming to produce visually coherent and high-fidelity outputs.

\begin{figure}[H]
    \centering
    \includegraphics[width=0.9\linewidth]{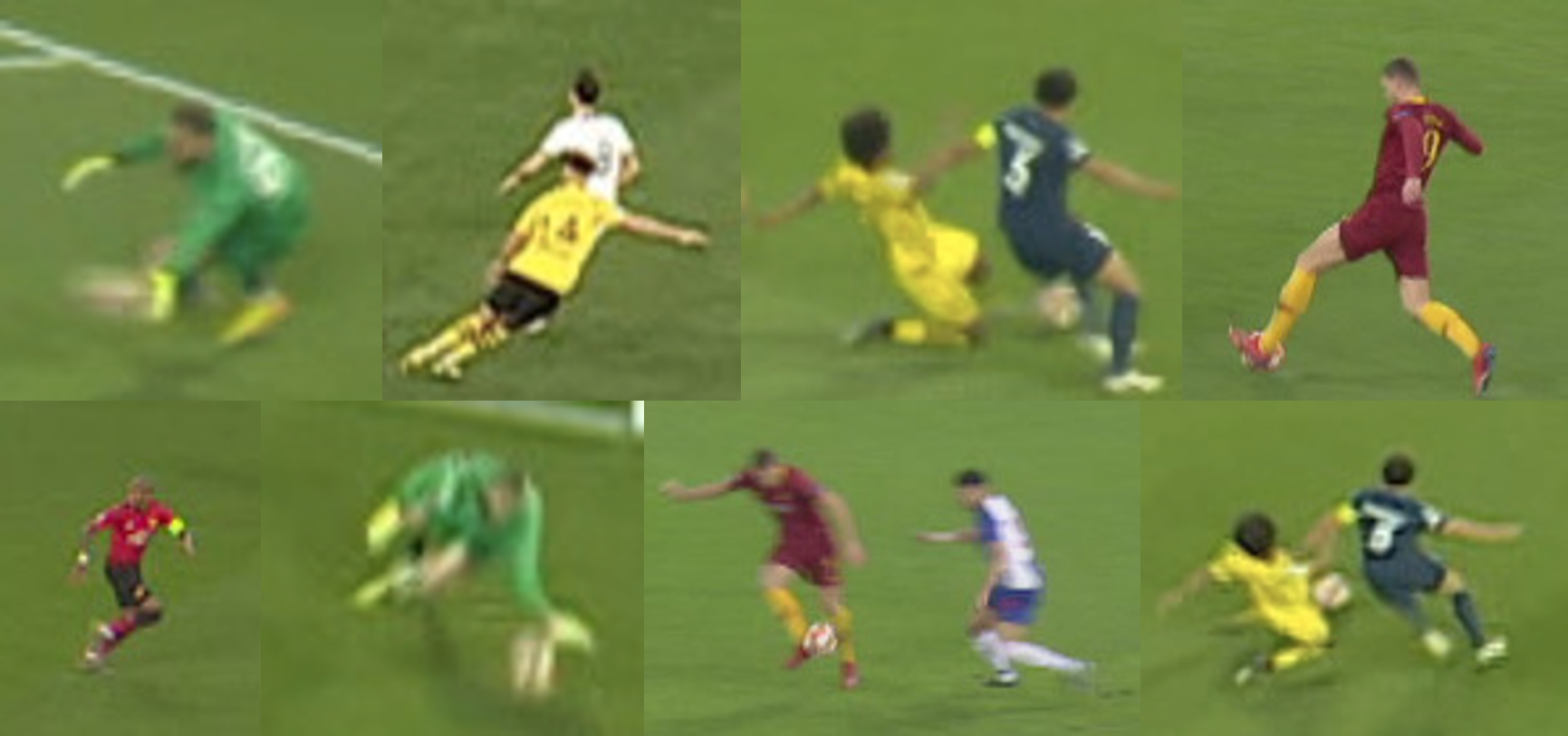}
    \caption{An example of raw image frames segmented from a 1080p interlaced broadcast. These snapshots, manually cut by technical operators during the match, exhibit varying resolutions, often starting as low as $60 \times 60$ pixels.}
    \label{fig1}
\end{figure}

The images targeted for upscaling are low-resolution snapshots originating from video production systems commonly used in broadcasting football events. These systems are designed to capture and process live video frames, segmenting them into individual images for transmission to operators. The transmitted images are typically used for tasks such as post-match analysis and reporting. Due to the interlaced nature of 1080p broadcasting, the segmented images are often significantly reduced in size. The snapshots start at an average resolution of around $100 \times 100$ pixels, as shown in Figure \ref{fig1}, presenting a considerable challenge in terms of restoring both clarity and missing details effectively.

Python 3.11 was selected as the programming environment, integrating PyTorch 2.5.1 and CUDA 12.4 to leverage GPU-accelerated computations. The Diffusers library \cite{huggingface_diffusers} was employed to construct and manage diffusion pipelines, offering fine-grained control and debugging capabilities superior to graphical interfaces like ComfyUI \cite{comfyanonymous_comfyui}. This modularity facilitated the customization of diffusion processes to meet the specific requirements of restoring and reconstructing broadcast frames.

The workflow was executed using an Nvidia RTX 6000 Ada GPU, which provided the necessary VRAM capacity to handle the computational demands of both training and inference. To monitor hardware performance and optimize resource utilization, the nvidia-ml-py3 \cite{nvidia_ml_py} library was employed, enabling real-time tracking of GPU metrics such as memory usage, temperature, and power consumption.

\begin{figure}[H]
    \centering
    \includegraphics[width=0.65\linewidth]{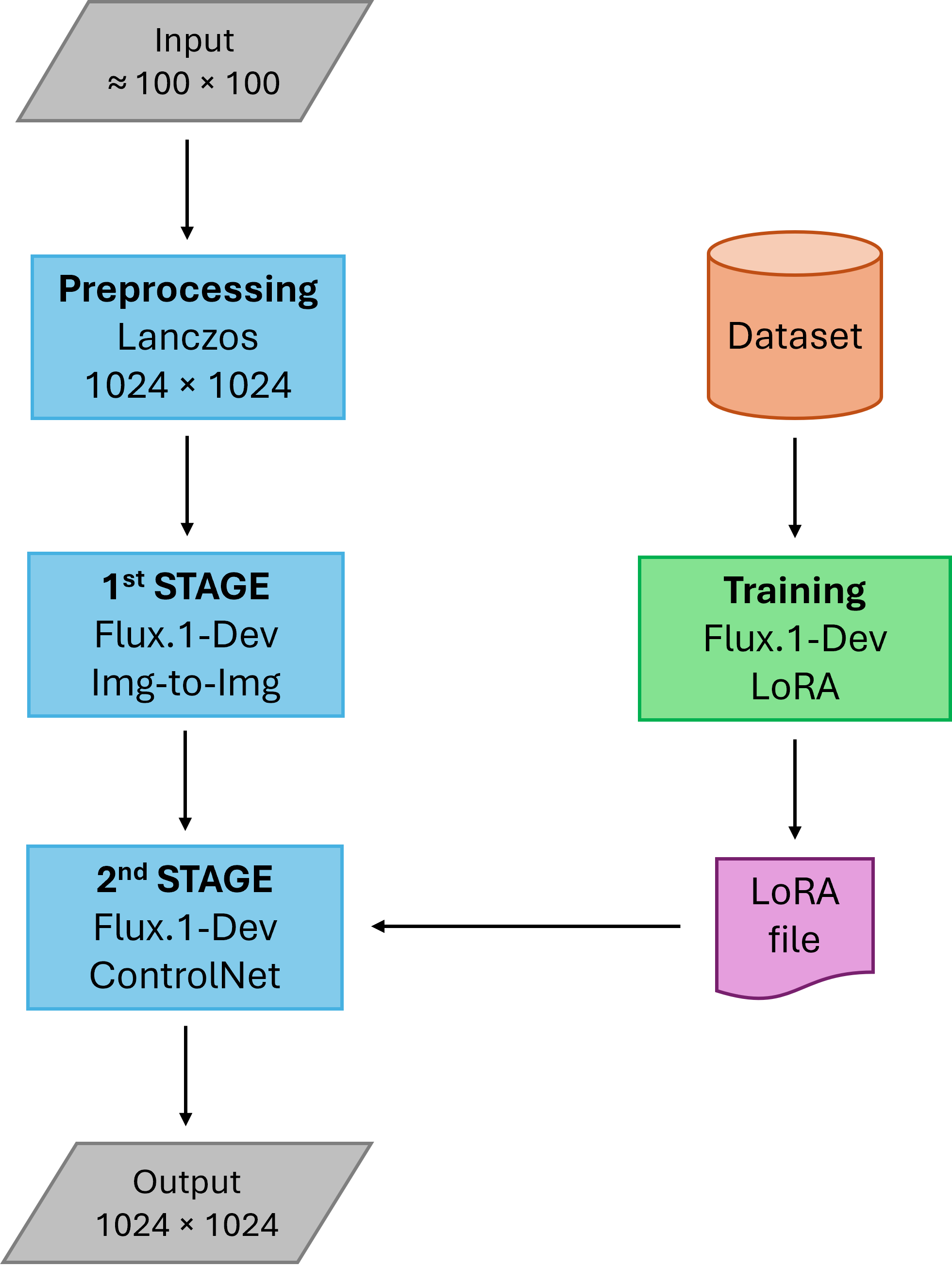}
    \caption{Complete pipeline flowchart}
    \label{fig2}
\end{figure}

The complete pipeline is presented in Figure \ref{fig2}, with each stage of the process discussed in its respective section.

\subsection{First stage of the reconstruction}\label{subsec2}

Before the actual reconstruction, the first phase focuses on standardizing the input images to ensure consistent dimensions and aspect ratios, addressing the variability inherent in the images extracted from live sports broadcasts. This variability, if left unmanaged, can lead to inconsistencies when using models optimized for specific input dimensions. Our ultimate objective was to reconstruct these images to a resolution of $1024 \times 1024$, not merely upscaling the images but also restoring critical details such as textures, logos and intricate visual elements that were lost or degraded in the original low-resolution inputs.

To achieve this, a preprocessing step was implemented to resize all input images to a square aspect ratio. The choice of a square format was deliberate, as many Diffusion Models are optimized for square inputs. Non-square images could introduce unintended variations, potentially confounding the outcomes when switching between models. Standardizing the aspect ratio ensures that results remain consistent across different experiments and model configurations.

The resolution for this preprocessing stage was set at $256 \times 256$ pixels, balancing computational efficiency and the retention of visual details. The resizing was performed using the Lanczos resampling algorithm \cite{pillow_concept_filters}, a high-quality upscaling method renowned for its ability to preserve edge details while minimizing aliasing. This approach ensured that the resized images retained as much fidelity as possible from their original forms, providing a solid foundation for subsequent stages of image reconstruction and enhancement.

\begin{figure}[H]
    \centering
    \includegraphics[width=0.9\linewidth]{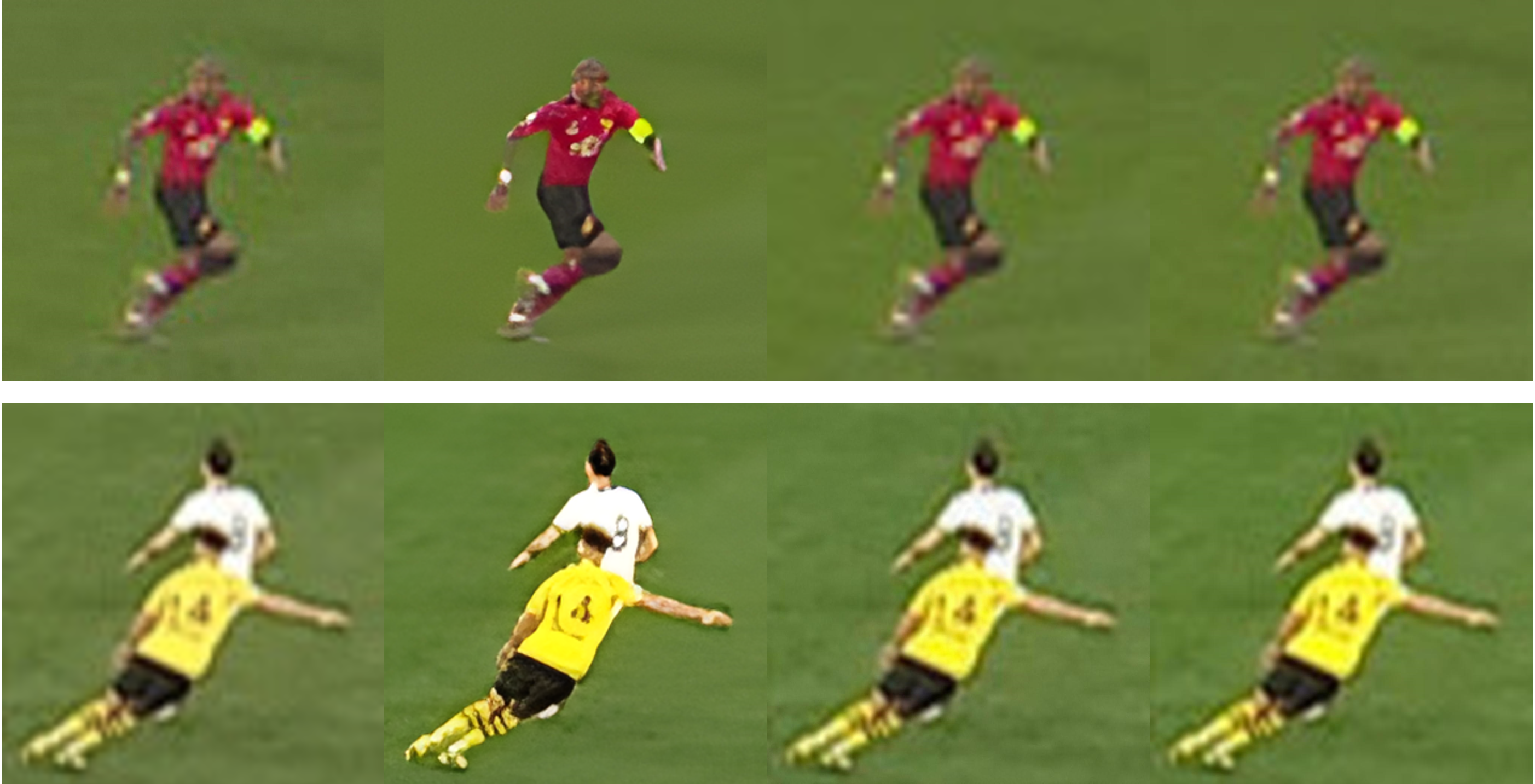}
    \caption{Comparison of image enhancement techniques applied to raw broadcast frames. From left to right, each column represents: the raw image frames resized to $256 \times 256$ using Lanczos resampling, the result of applying Real-ESRGAN for $1024 \times 1024$ upscaling, the result from the latent x4 upscaler (Stable Diffusion x4 Upscaler), and finally, the Lanczos resampling upscaled to $1024 \times 1024$.}
    \label{fig3}
\end{figure}

Initially, to achieve the reconstruction, we attempted to use the Real-ESRGAN algorithm, leveraging the Upscayl application \cite{upscayl} for implementation. Upscayl is an open-source, user-friendly tool designed for upscaling and enhancing image resolution using cutting-edge algorithms like Real-ESRGAN. It provides a streamlined interface for leveraging advanced AI models without requiring in-depth technical expertise, making it a popular choice for quick image enhancement tasks. However, as shown in Figure \ref{fig3}, the results were extremely poor. This outcome is likely due to the significant degradation present in the input images, which exceeded Real-ESRGAN's ability to reconstruct satisfactory details. The severe blurring, noise, and compression artifacts in our dataset highlighted the limitations of Real-ESRGAN for this specific application. This prompted us to explore alternative approaches to achieve the desired quality and resolution.

After encountering the limitations of Real-ESRGAN, we next tried using a latent upscaler to enhance the resolution of the degraded images. Specifically, we utilized the Stable Diffusion x4 Upscaler model available in the Diffusers library and developed by StabilityAI \cite{stabilityai_upscalerx4}. This model is a specialized variant of Stable Diffusion, designed to upscale images by a factor of four while leveraging latent diffusion techniques. Unlike traditional pixel-space upscaling methods, this model operates in the latent space, capturing high-level semantic details before decoding them into high-resolution outputs. The latent diffusion process allows for the generation of more coherent and visually appealing results, especially in scenarios with moderate degradation.
While the results were slightly better compared to Real-ESRGAN, the upscaled images were still heavily degraded and remained largely unusable for our purposes, as shown in Figure \ref{fig3}. The pervasive blurring, lack of fine detail, and persistent artifacts highlighted the challenges of applying such techniques to our highly degraded dataset, further emphasizing the need for a more robust and domain-specific reconstruction approach. 

Given that the previous algorithms, both Real-ESRGAN and the Stable Diffusion x4 Upscaler, failed to deliver significant advantages in detail reconstruction (Figure \ref{fig3}) and introduced considerable computational overhead, Lanczos resampling emerged as a far more efficient and practical solution. Unlike neural network-based methods, which involve complex architectures and substantial computation, Lanczos relies on a straightforward mathematical resampling technique. This simplicity not only makes it faster and less resource-intensive but also ensures consistent and reliable upscaling results. Its ease of computation and efficiency make Lanczos an ideal choice for standardizing input images before applying more advanced diffusion-based enhancement techniques.

After evaluating the limitations of the previously tested methods, we decided to adopt an image-to-image pipeline approach, which is widely recognized in the Diffusers library. This approach leverages the inherent strengths of Diffusion Models in iterative refinement, allowing us to transform low-resolution and degraded images into high-resolution outputs while reconstructing fine details. Since we aimed to evaluate the latest SOTA models, we upscaled the images to 1024×1024 resolution using Lanczos resampling: most of Diffusion Models are optimized for this resolution since are trained on images of this size.

Following the preprocessing step with Lanczos resampling, we began experimenting with various Diffusion Models using the Diffusers library \cite{huggingface_diffusers} to develop an effective image-to-image reconstruction pipeline. This method leverages the power of Diffusion Models to enhance input images by iteratively refining noise while maintaining the structural coherence of the original image. Unlike generative processes that synthesize images from scratch, the image-to-image pipeline conditions the denoising process on the input image, ensuring that the enhanced output aligns with the original context and layout.

Technically, the pipeline works by first adding a controlled amount of noise to the input image. This noising step, parameterized by a \texttt{strength} argument (Code \ref{code1}), determines the extent of noise introduced into the input image and, as a result, the degree of deviation allowed during the enhancement process. \texttt{Strength} is one of the most critical parameters in the pipeline, as it directly impacts the generated image \cite{huggingface_diffusers_img2img}. Specifically:

\begin{itemize}
    \item A higher \texttt{strength} value gives the model greater ``creativity'', allowing it to produce an output that is increasingly different from the initial image. At a \texttt{strength} value of 1.0, the initial image is effectively ignored and the generation is only based on the textual prompt.
    \item A lower \texttt{strength} value means the generated image closely resembles the original input, prioritizing fidelity over transformation.
\end{itemize}
 
The \texttt{strength} and \texttt{num\_inference\_steps} (Code \ref{code1}) parameters are closely related, as \texttt{strength} determines how many noise steps are applied to the input image. For example, with \texttt{num\_inference\_steps} set to 50 and \texttt{strength} set to 0.8, the process involves adding 40 noise steps ($50 \times 0.8$) to the input image, followed by 40 steps of denoising to produce the output. This relationship underscores the need to carefully balance these parameters for optimal results. During the denoising phase, the model iteratively removes noise at each step, using the conditioning input (the original image) and optional prompts or textual descriptions as guidance. This step-by-step refinement not only restores missing details but also enhances the image with sharper textures and greater visual intricacy.

In addition to \texttt{strength} and \texttt{num\_inference\_steps}, other parameters (Code \ref{code1}) play a crucial role in controlling the behavior and output of the pipeline: 
\begin{itemize}
    \item The \texttt{seed} parameter is used to initialize the random number generator, ensuring reproducibility of results. By setting a specific seed value (e.g., \texttt{seed = 5}), the same noise pattern is generated for each run. This consistency is particularly useful for comparing different configurations or testing enhancements. The \texttt{generator} is a PyTorch object initialized with the manual seed, providing fine-grained control over the random sampling process.
    \item The \texttt{prompt} parameter defines the textual description used to guide the diffusion process. 
    \item The \texttt{guidance\_scale} controls the influence of the prompt on the generated image. Higher values prioritize adherence to the prompt at the cost of potentially reduced coherence with the original image, while lower values allow the model more freedom to balance the prompt and the input image.
\end{itemize}

\renewcommand{\lstlistingname}{Code}
\lstset{
    breaklines=true,           
    breakatwhitespace=false,   
    basicstyle=\ttfamily\small, 
    numbers=none,              
    frame=single,              
    backgroundcolor=\color{gray!10}, 
    xleftmargin=0.5cm,        
}
\begin{lstlisting}[float, language=Python, caption={Example of image-to-image generation pipeline with Diffusers library}, label={code1}]
from PIL import Image
from diffusers import AutoPipelineForImage2Image

pipeline = AutoPipelineForImage2Image.from_pretrained(
                "stabilityai/stable-diffusion-3.5-medium", 
                ).to("cuda")

seed = torch.random.seed()
generator = torch.Generator().manual_seed(seed)
prompt = ("A dog")
image = pipeline(
    prompt=prompt,
    image=Image.open("img.jpg"),
    guidance_scale=3.5,
    strength=0.6,
    height=1024,
    width=1024,
    num_inference_steps=90,
    generator=generator
).images[0]

image.save("dog.jpg")
\end{lstlisting}

These parameters, when tuned carefully alongside \texttt{strength} and \texttt{num\_inference\_steps}, enable precise control over the pipeline, ensuring high-quality reconstructions that align with both the input image and the descriptive prompt.

\begin{figure}[H]
    \centering
    \includegraphics[width=0.95\linewidth]{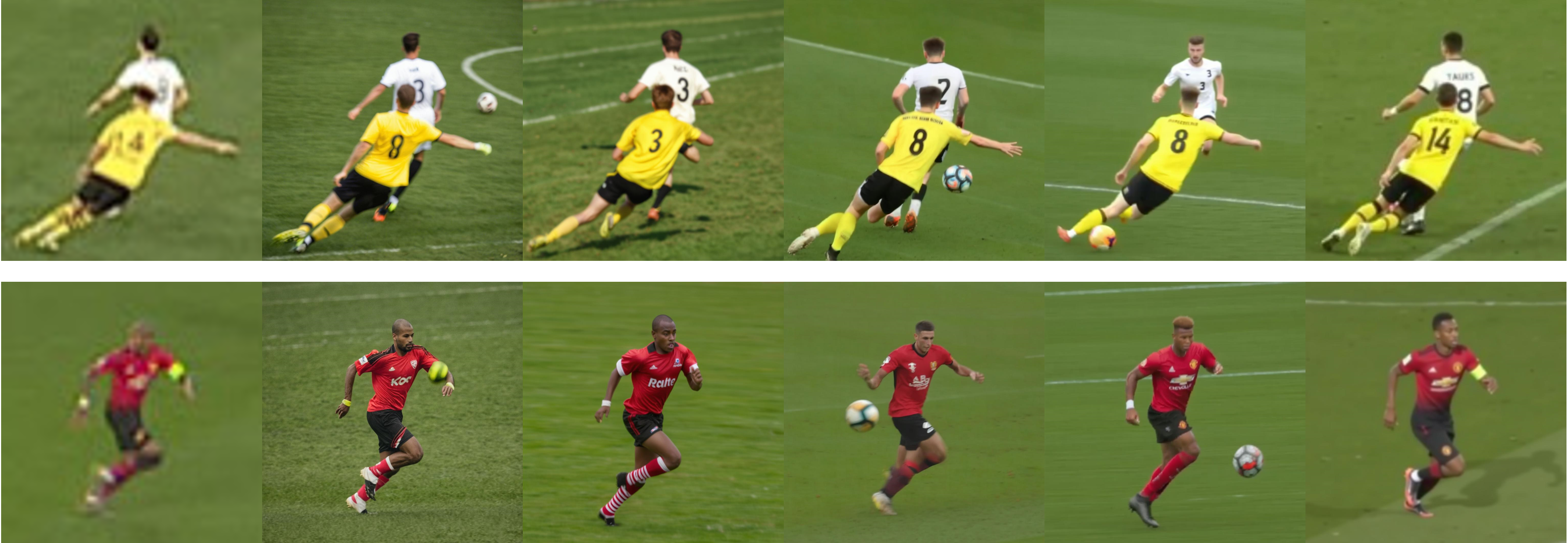}
    \caption{Comparison of outputs from image-to-image pipelines using various models. From left to right, each coloumn represents: the original raw frames, the outputs of Stable Diffusion XL, RealVis\_v4.0, Stable Diffusion 3.5-Medium, Stable Diffusion 3.5-Large, and FLUX.1-dev.}
    \label{fig4}
\end{figure}

Initially, we explored a range of open-source models to tackle the challenges of reconstructing fine-grained details in highly degraded images. This included Stable Diffusion XL \cite{stabilityai_stable_diffusion_xl_base}, the most recent Stable Diffusion 3.5 in both \textit{medium} \cite{stabilityai_stable_diffusion_3_5_medium} and \textit{large} \cite{stabilityai_stable_diffusion_3_5_large} configurations, as well as several fine-tuned models designed to produce realistic outputs. Notable among these were models available on platforms such as CivitAI \cite{civitai_website}, including RealVis\_V4.0 \cite{sg161222_realvisxl_v4}. Although these models exhibited significant strengths in general applications, they fell short in meeting the specific requirements of extreme upscaling and intricate detail recovery within the football domain, as shown in Figure \ref{fig4}. 

After evaluating various open-source models, we selected FLUX.1-dev \cite{black_forest_labs_flux1_dev_HF} as our primary model due to its superior performance in our football-specific context. FLUX.1-dev is an open-weight, guidance-distilled model designed for non-commercial applications. It is directly distilled from FLUX.1-pro, achieving similar quality and prompt adherence capabilities while being more efficient than standard models of the same size \cite{black_forest_labs_website}. The FLUX.1 architecture is based on a hybrid design that combines multimodal and parallel diffusion transformer blocks, scaled to 12 billion parameters. This structure enhances the model's ability to generate high-quality images with detailed textures and complex visual elements.

We observed superior performance with FLUX.1-dev compared to other models, particularly against Stable Diffusion 3.5 which is particularly surprising given that Stable Diffusion 3.5, as one of StabilityAI's latest models, should be expected to compete closely with FLUX.1-dev. Although Stable Diffusion 3.5 utilizes three Text Encoders, two of these, CLIP-L/14 \cite{openai_clip_vit_large_patch14} and T5XXL \cite{google_t5_v1_1_xxl}, are also incorporated in FLUX.1-dev. Both models share a DiT backbone \cite{stabilityai_stable_diffusion_3_5_medium} \cite{huggingface_diffusers_fluxAPI} and leverage the Rectified Flow method to enhance training efficiency and stability in the diffusion process. Despite these architectural similarities, Stable Diffusion 3.5 delivered far inferior results in most of our experiments, failing to achieve the same level of detail recovery and visual coherence. FLUX.1-dev consistently excelled in reconstructing fine details and intricate textures in our specific domain, making it the clear choice for our workflow. Additionally, FLUX.1-dev demonstrated significantly better coherence with the original image, even when used with high \texttt{strength} values, avoiding excessive deviation from the input. Moreover, FLUX.1-dev exhibits superior prompt adherence and enhanced reconstruction capabilities for company brands and logos, a feature crucial for applications in a broadcast company.

However, this high-quality output comes at the expense of increased computational demands. Running the full FLUX.1-dev model with both text encoders active requires a minimum of 24 GB of VRAM, highlighting the model's resource-intensive nature. Additionally, running FLUX.1-dev on consumer-grade hardware, even with 24 GB of VRAM, can be challenging without optimizations such as CPU offloading and memory-efficient computation techniques \cite{hf_cpuoff_sd3_memory_optimizations}. These optimizations help manage the substantial memory requirements and computational load, making it feasible to deploy the model in environments with limited resources. To address these challenges, we utilized the following optimization techniques: 

\begin{itemize}
    \item Bfloat16 precision: This format, supported by modern GPUs and CPUs, reduces memory usage by using 16-bit floating-point precision while retaining the dynamic range of 32-bit floats. This allows for faster computation and reduced memory requirements without a significant impact on the quality of the outputs \cite{huggingface_fast_diffusion_acc_inference}.
    \item Model CPU offloading: This feature offloads specific model components, such as the attention blocks and text encoders, to the CPU during both training and inference. By leveraging the memory bandwidth of the CPU, this approach reduces the GPU memory burden and enables running large models like FLUX.1-dev on hardware with limited VRAM. Additionally, it ensures smooth operation by transferring data between the GPU and CPU dynamically during processing \cite{hf_cpuoff_sd3_memory_optimizations}.
\end{itemize}

Our workflow begins with Lanczos resampling (\texttt{resize\_image\_square} function in the Code \ref{code2}), a precise and efficient method for resizing/upscaling low-resolution inputs to the required $1024 \times 1024$ resolution for processing within the Flux.1-dev framework. This resampling stage is critical, as it preserves essential details while mitigating artifacts typically introduced by other interpolation methods, thereby creating a clean and reliable baseline for downstream processing.

In the next stage, a Flux.1-dev image-to-image pipeline was configured with a \texttt{strength} parameter set at 0.75 and \texttt{num\_inference\_steps} fixed at 80 (Code \ref{code2}). These values were meticulously calibrated to strike an optimal balance between maintaining structural integrity and enhancing fine details. The pipeline utilized a carefully crafted prompt, tailored to align with the characteristics of the input images and the desired output quality. Additionally, to ensure consistency, a fixed \texttt{seed} was employed throughout the process, enabling systematic exploration of configurations and facilitating the identification of the best settings for the task. However, in some cases, we kept the \texttt{seed} random and used the \texttt{num\_images\_per\_prompt} parameter to generate multiple images at once, allowing us to evaluate the outputs and select the most suitable one for our needs.

At this developmental stage, prompts were manually crafted, leveraging information provided by the broadcast technician about specific elements within the scene, such as numbers on players’ shirts. This knowledge proved invaluable in enhancing the accuracy and coherence of the generated outputs. For instance, an example prompt used was: ``A goalkeeper wearing a green jersey and matching shorts with the number 99 is diving low to make a save. The player is focused on controlling the ball, which is on the ground ahead of him. The scene unfolds on a soccer field, with the blurred background featuring the goal area and field markings''.

The \texttt{guidance\_scale} parameter was set to 3.5, as it proved to deliver the best results in terms of prompt adherence, ensuring the outputs remained aligned with the intended characteristics defined by the crafted prompts.

Despite leveraging the Nvidia RTX 6000 Ada GPU with 48 GB of VRAM for inference, \texttt{enable\_model\_cpu\_offload} were employed to circumvent memory exhaustion and maintain stability. This optimization ensured sufficient memory availability for subsequent stages of reconstruction. Furthermore, the use of components like the AutoEncoder, two Text Encoders, and DiT in bfloat16 format significantly accelerated the inference process while maintaining computational efficiency and improving latency.

\begin{lstlisting}[float, language=Python, caption={Our pipeline used for the first image reconstruction stage}, label={code2}]
import torch
from diffusers import FluxImg2ImgPipeline
from PIL import Image

pipe = FluxImg2ImgPipeline.from_pretrained(
                            "black-forest-labs/FLUX.1-dev", 
                            torch_dtype=torch.bfloat16
                            )
pipe.enable_model_cpu_offload() # entire model is offloaded

seed=5
generator=torch.Generator().manual_seed(seed)
prompt= # prompt changes based on the input image
image=pipe(
    prompt=prompt,
    image=resize_image_square("imgs/99 Donnarumma.jpg", 1024), 
    # custom function for Lanczos resampling
    num_images_per_prompt=3
    guidance_scale=3.5,
    strength=0.75,
    height=1024,
    width=1024,
    num_inference_steps=80,
    generator=generator
).images
\end{lstlisting}

\subsection{Second stage of the reconstruction}\label{subsec3}

After completing the initial reconstruction with an image-to-image pipeline (Figure \ref{fig2}), we achieved significantly enhanced $1024 \times 1024$ images compared to the degraded originals. This stage successfully restored substantial amounts of detail and structure, elevating the visual quality of the outputs. However, a consistent challenge emerged during these experiments: many reconstructed images exhibited a noticeable degree of blurriness.

The primary cause of this issue was the need to constrain the model’s deviation from the original input. To achieve this, the \texttt{strength} parameter in the image-to-image pipeline was kept below 0.8. While this precaution was crucial for preserving the overall structure and critical elements of the input images, it also limited the model’s ability to introduce finer details or sharpen features effectively. Consequently, the resulting images, though significantly improved, often lacked the crispness and clarity required for high-fidelity applications such as sports broadcast analysis.

To address the limitations of the first stage, the second stage of the reconstruction process leverages ControlNet, a conditional refinement network designed to provide precise control over the generative process. By introducing auxiliary conditioning inputs, ControlNet enhances the model’s capacity to refine specific aspects of the image while maintaining adherence to its original context. 

In this stage, we employed a pre-trained ControlNet model, Flux.1-dev-ControlNet-Upscaler \cite{Jasperai/flux.1-dev-controlnet-upscaler}. The inputs to this model were the reconstructed $1024 \times 1024$ outputs from the first stage. This ControlNet model was trained using a synthetic complex data degradation scheme that involved artificially degrading real-life images by combining several degradation techniques, including Gaussian and Poisson noise, image blurring, and JPEG compression, in a manner similar to Real-ESRGAN \cite{RealESRGAN2021}. The \texttt{controlnet\_conditioning\_scale} parameter (Code \ref{code3}) in ControlNet pipelines determines the influence of ControlNet’s output on the base U-Net or DiT model during image generation. Specifically, the outputs from ControlNet are multiplied by this scale factor before being added to the residual connections in the U-Net or DiT. Adjusting this parameter allows control over how strongly the conditioning input affects the final generated image \cite{huggingface_controlnet_api}.

\begin{lstlisting}[float, language=Python, caption={Pipeline used for the second image reconstruction stage}, label={code3}]
import torch
from diffusers import FluxControlNetModel
from diffusers.pipelines import FluxControlNetPipeline

controlnet = FluxControlNetModel.from_pretrained(
                  "jasperai/Flux.1-dev-Controlnet-Upscaler",
                  torch_dtype=torch.bfloat16)
pipe = FluxControlNetPipeline.from_pretrained(
                  "black-forest-labs/FLUX.1-dev",
                  controlnet=controlnet,
                  torch_dtype=torch.bfloat16)

pipe.enable_model_cpu_offload()

control_image = Image.open("reconstructed1024.jpg")
seed = torch.random.seed()
generator = torch.Generator().manual_seed(seed)
image = pipe(
    prompt= # prompt changes based on the input image,
    control_image=control_image,
    num_images_per_prompt=3
    controlnet_conditioning_scale=0.5,
    num_inference_steps=35, 
    guidance_scale=3.5,
    height=1024,
    width=1024,
    generator = generator
).images
\end{lstlisting}

\begin{figure}[h]
    \centering
    \includegraphics[width=0.75\linewidth]{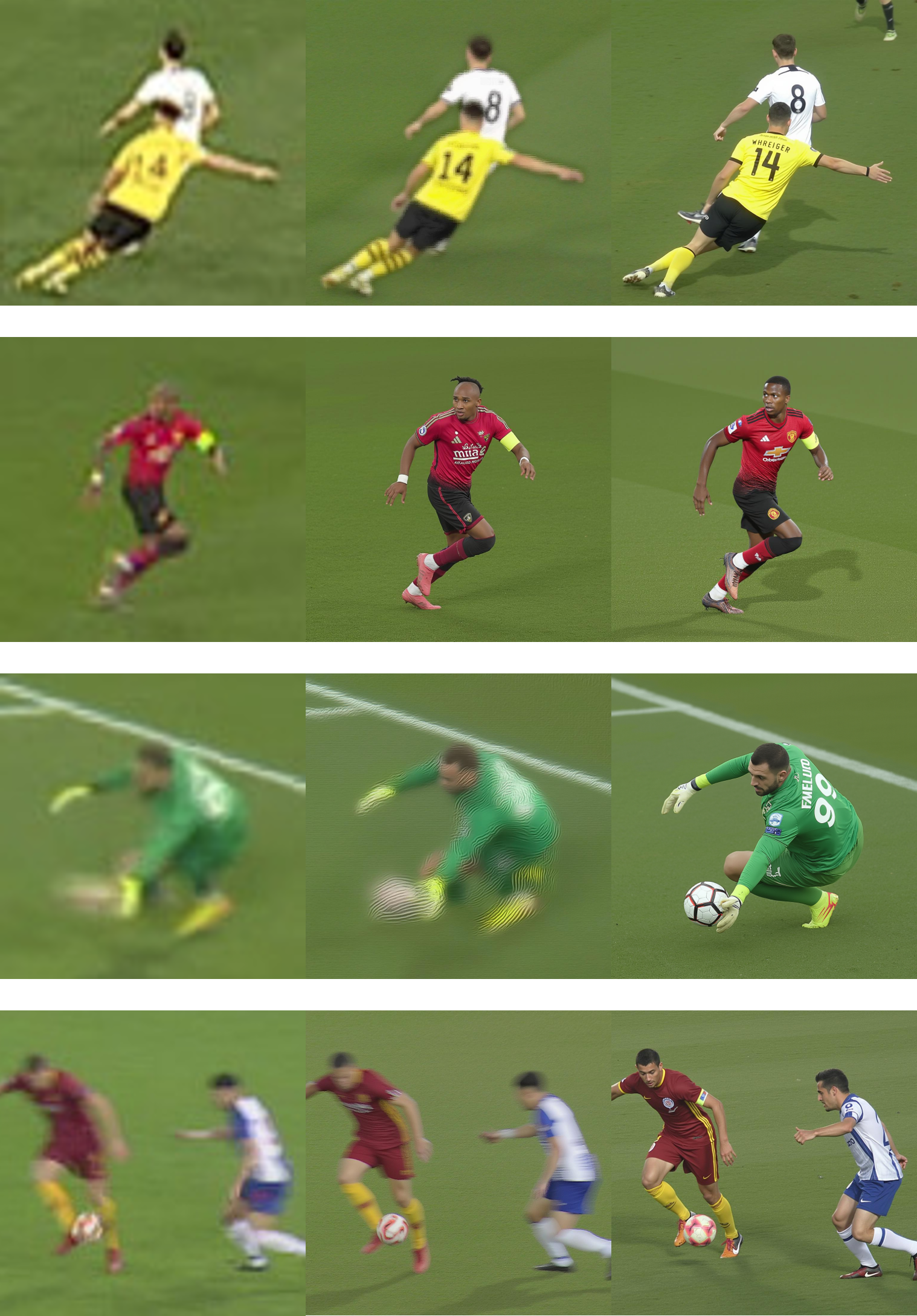}
    \caption{Comparison of Control-Net pipeline outputs. From left to right, each coloumn represents: the original raw frames, the output images without the first stage of the reconstruction, the output images with the first stage of the reconstruction.}
    \label{fig5}
\end{figure}

The incorporation of ControlNet enabled us to apply fine-grained modifications, targeting areas where detail restoration was most needed. For example, ControlNet effectively sharpened blurred player numbers, refined jersey textures, and restored field markings while preserving the broader composition of the image, as shown in Figure \ref{fig5}. The pipeline was optimized by setting the \texttt{controlnet\_conditioning\_scale} parameter between 0.5 and 0.65, ensuring a balanced influence of the conditioning input relative to the textual prompt. The textual prompts used in this stage were identical to those used in the first stage.

However, this enhancement comes at the cost of increased computational requirements. Integrating ControlNet with the Flux.1-Dev pipeline incurs an additional VRAM overhead of approximately 4 GB, bringing the total VRAM usage to 30 GB for generating a single image. This high memory demand necessitates the use of hardware with sufficient VRAM capacity for stable and efficient execution.

Furthermore, it is crucial to point out that the first stage of reconstruction was not only instrumental in improving the degraded images but also fundamental for the success of the subsequent ControlNet pipeline. When the raw image frames were fed directly into the ControlNet pipeline, the results were extremely poor (Figure \ref{fig5}), as the severe degradation in the original frames overwhelmed the model’s capacity for conditional refinement. The initial reconstruction provided a critical baseline by reducing noise, enhancing structure, and restoring key elements, thereby creating a reliable foundation for ControlNet to build upon.

By leveraging ControlNet in this second stage, we successfully overcame the limitations of the initial reconstruction pipeline. This multi-stage framework, which combines the strengths of image-to-image pipelines with conditional refinement, provides a robust solution for reconstructing and enhancing degraded images in sports broadcasting contexts.

\clearpage
\subsection{Fine-tuning}\label{subsec4}

The results achieved so far using the multistage framework, which combines an image-to-image pipeline and ControlNet, have been satisfactory in most cases. However, we believed there was room for further improvement in the final outputs. To address this, we decided to train a LoRA for the FLUX.1-Dev model, aiming to enhance its reconstruction capabilities specifically for football actions such as interceptions, headers, or sliding tackles, as well as details like shirt logos and player skin tones.

A key aspect of this work is the custom dataset we developed for training. Since we could not find a proper football dataset online with high-quality images paired with captions, we created one ourselves. The dataset comprises high-resolution images paired with descriptive captions, and all the images were sourced internally from the broadcast company we collaborate with, ensuring both quality and specificity to the task.

For training, we utilized two RTX 6000 Ada GPUs to accelerate the process and relied on the KohyaSS repository \cite{kohya_sd_scripts}. This repository offers highly customizable Python scripts tailored for fine-tuning various Diffusion Models, enabling precise adjustments to achieve optimal results.

\subsubsection{Dataset}\label{subsubsec2}

The broadcast company provided us with 460 high-quality football images, which will undergo various augmentation techniques to achieve a satisfactory number before training. To accelerate the dataset creation process, we decided to use ChatGPT Plus \cite{openai_website}, specifically creating a custom GPT, tailored to generate detailed captions for the included images. This custom GPT was designed to provide rich descriptions of key elements commonly observed in football-related imagery.

The primary goal of the GPT was to generate precise captions capable of explaining the depicted game elements, actions, and the players involved. The sole input required from the user during the image upload process was identifying the teams present in the image, which was necessary when describing them for the first time. The GPT  was designed with a specific set of instructions to ensure the generation of detailed, accurate, and professional captions for soccer-related images. Its primary objective was to produce captions that adhered to strict guidelines, guaranteeing consistency and clarity across all outputs. The rules emphasized the need for descriptive and neutral language, restricting the captions to observable elements within the image to avoid subjective interpretations or assumptions.

Captions were required to begin with a precise identification of all visible individuals within the image. The GPT was instructed to provide an accurate count of these individuals while clearly distinguishing their roles, such as players, referees, coaches, or spectators, whenever possible. Specific attention was given to players' uniforms, ensuring detailed descriptions of their appearance. When players wore their team’s first jersey (home kit), the team name was explicitly mentioned in the caption. For second or third kits, team names were deliberately omitted to avoid ambiguity.

Descriptions of uniforms included a detailed breakdown of jersey and shorts colors, allowing for visual clarity and consistency. For instance, captions highlighted combinations such as ``red jersey with white shorts'' or ``blue jersey with yellow accents''. Uniform-specific details were dynamically associated with the respective team within the GPT chat when the user provided this information during the image upload process. Allowing the user to input the relevant context at the time of upload reduced the significant time burden of predefining all potential jersey-and-shorts combinations for every team. The process also accommodated the variability in secondary kits, which often change from season to season. Efficiency and adaptability were thereby ensured while maintaining the accuracy and relevance of captions.

Jersey numbers were included in the caption only when fully visible in the image. The GPT was instructed to link jersey numbers to specific actions performed by players, such as ``Player number 9 is kicking the ball''. Names of players were added only if clearly visible, ensuring captions relied exclusively on observable and unambiguous information. This approach allowed the GPT to learn the association between jersey numbers and specific player actions effectively.

Players' appearances were meticulously described, with hairstyles categorized into predefined groups such as bald, short, or long hair to ensure consistency across captions. Unique features, including braids, ponytails, or other distinctive arrangements, were incorporated when clearly visible, enhancing both the richness and accuracy of the descriptions. The GPT was explicitly instructed to provide descriptions that avoided any culturally biased or subjective terminology, focusing solely on observable features. Skin tones were described using neutral and professional language, emphasizing an observational and objective approach. The categorization avoided overly simplistic labels and focused on respectful descriptors that accounted for the diversity of individuals depicted in the images. Such guidelines were implemented to ensure that the captions remained precise, inclusive, and free from any unintended biases, aligning with the overall objective of creating a comprehensive and fair representation of the individuals in the dataset.

Instructing the GPT to adhere to these principles not only enhanced the quality of the captions but also reinforced the importance of ethical considerations in automated descriptive tasks. Every effort was made to ensure that descriptions contributed to an accurate yet impartial portrayal of players, avoiding any potential misinterpretation or cultural insensitivity.

The background of the image was described with attention to the stadium's occupancy and other visible elements. Captions specified whether the stadium was empty (no spectators), half full (moderate crowd), or at full capacity (sold-out stadium). The inclusion of these details provided additional context regarding the atmosphere and scale of the event. Observations also extended to identifying prominent background elements such as billboards, goalposts, or the corner flag whenever they were visible in the image. Providing these contextual elements contributed to a richer and more complete description of the scene, ensuring that the captions captured not only the primary action but also the environment in which it occurred.

The actions and movements of players were described in detail, focusing on activities such as running, kicking, tackling, celebrating, or interacting with the ball or other players. Captions provided clear and specific descriptions, for example: ``Two players are contesting possession of the ball in a sliding tackle''. Spatial relationships between players, objects, and field landmarks were explicitly outlined to enhance clarity, such as ``Player number 10 is positioned near the penalty box''. Consistency in terminology was emphasized, ensuring that similar actions were described using the same phrases across different captions. This standardization was critical for training the GPT to recognize and accurately describe recurring patterns of play.

Uniform descriptions were mandatory for all visible players, detailing the colors of both jerseys and shorts worn by each team. Captions included statements like ``The player is wearing a blue jersey with white shorts'' or ``The red team is in a red jersey and red shorts, while the opposing team is wearing a white jersey with black shorts''. Such precision provided consistency and allowed for clear differentiation between teams, contributing to the overall comprehensiveness of the captions.

Descriptions focused exclusively on elements that were clearly visible, ensuring accuracy and reliability. Sentences were crafted to be concise, grammatically correct, and written in clear English, prioritizing clarity and readability. These guidelines supported the creation of high-quality, objective captions that aligned with the project's goals for precision and consistency.

The captions were required to be formatted in a single JSON file by the KohyaSS scripts, where each image file path was directly linked to its respective caption (Code \ref{code4}). This design ensured seamless integration into downstream applications or analyses while maintaining high-quality descriptive standards. Captions were required to be accurate, objective, and formatted correctly for immediate use. By embedding the image paths as keys in the JSON structure, the need for an image\_name field was eliminated, avoiding redundancy. This approach streamlined the process, ensuring that each caption was directly associated with its corresponding image path, thereby enhancing the overall efficiency and usability of the dataset.

\begin{lstlisting}[float, caption={Example of the structured JSON file with captions describing the actions, appearance, and context of players and the stadium in a soccer image.}, label={code4}]
{
    "/workspace/imgs/imago1050756553.jpg": {
        "caption": "A player from A.C. Milan, in a red and black striped jersey with white shorts, marked with the number 11, is attempting to score as he approaches the goal. He is surrounded by two players from Inter Milan, in black and blue striped jerseys, one with the number 28 and another with the number 20, trying to block him. The goalkeeper from Inter Milan, wearing a black kit, is preparing to make a save. The background shows the net, advertising boards, and a partially filled stadium with spectators."
    },
    "/workspace/imgs/imago1050756556.jpg": {
        "caption": "A player from Inter Milan, with short hair and white skin tone, in a black and blue striped jersey with black shorts, is in possession of the ball while being challenged by a player from A.C. Milan, with curly hair and white skin tone, in a red and black striped jersey with white shorts. Another A.C. Milan player, with short hair and dark skin tone, also in a red and black striped jersey, is positioned nearby, observing the play. The action is taking place on a grass field, with advertising boards and out-of-focus spectators in the background."
    },
    # ...
}
\end{lstlisting}

\clearpage

\subsubsection{Training}\label{subsubsec3}
For training the LoRA model on our football dataset, we leveraged the KohyaSS \cite{kohya_sd_scripts} training pipeline: this was chosen due to its support for fine-tuning large-scale Diffusion Models and its flexibility in accommodating advanced training strategies.

The Kohya pipeline expects the dataset configuration to be defined in a separate .toml file (Code \ref{code5}), where a variety of configuration parameters can be set \cite{kohya_dataset}. This structured approach enables precise control over the dataset and its preparation.

\begin{lstlisting}[caption={.toml file for dataset configuration}, label={code5}]
[[datasets]]
resolution = [1024, 1024]
enable_bucket = true
max_bucket_reso = 1024

  [[datasets.subsets]]
  image_dir = '/workspace/imgs'
  metadata_file = '/workspace/captions_kohya.json'
  flip_aug = true
  num_repeats = 1
  shuffle_caption = false
\end{lstlisting}

The main parameters that are important and peculiar to point out are:
\begin{itemize} 
    \item \texttt{enable\_bucket}: This parameter activates the bucket-based resolution mechanism, which dynamically groups images into resolution ``buckets'' according to predefined sizes. This approach ensures efficient GPU memory utilization by resizing images to fit within the allocated buckets, leading to faster and more consistent training. The upper resolution limit for bucket allocation is determined by \texttt{max\_bucket\_reso}, meaning that images exceeding this resolution threshold are resized accordingly to remain within the defined constraints.
    \item \texttt{flip\_augmentation}: Enables horizontal flipping as a data augmentation technique. This method effectively doubles the dataset size by generating mirrored versions of images, which is particularly useful in the domain of football imagery. Since football actions exhibit left-to-right symmetry without altering the semantics of the scene, this augmentation enhances generalization without introducing inconsistencies.
    \item \texttt{num\_repeats}: Specifies the number of times each image-caption pair is repeated during training. This parameter ensures a balanced representation of underrepresented samples, such as specific football-related actions (e.g., headers, tackles), thereby improving the model's ability to generalize. Given that our dataset comprises only 460 images, we set this value to 5 to increase exposure to each image, thereby facilitating effective learning by the model.
\end{itemize}

The application of this optimizations and augmentations resulted in a dataset of 2300 images, coming from $460 \times 5$ repeats, which are randomly and horizontally flipped.

Since we used two Nvidia RTX 6000 Ada GPUs for training, to fully leverage the hardware capabilities the environment was configured using the Accelerate library \cite{huggingface_accelerate_docs} in a \texttt{multi\_gpu} distributed type setup. This enabled efficient distribution of computational tasks across both GPUs, optimizing memory usage and reducing overall training time.

During the configuration of \texttt{Accelerate}, we explored integrating the DeepSpeed library \cite{microsoft_deepspeed} to optimize memory efficiency further and enable advanced features such as ZeRO (Zero Redundancy Optimizer). This technique significantly reduces memory requirements by partitioning optimizer states, gradients, and model parameters across GPUs. Specifically, we experimented with ZeRO Stages 1, 2, and 3 to evaluate their compatibility with our setup. However, several challenges arose: Stages 2 and 3 were found to be incompatible with Kohya scripts, while Stage 1, despite being explicitly supported through the \texttt{deepspeed} argument, encountered a matrix multiplication issue within PyTorch, preventing successful execution. These limitations necessitated alternative optimization strategies for memory efficiency and scalability. Furthermore, at this time, fine-tuning scripts for LoRA remain highly experimental, with some implementations not even attempting to incorporate DeepSpeed support.

In addition, we attempted to utilize Torch Dynamo \cite{pytorch_compiler_dynamo} with both \texttt{'inductor'} and \texttt{'cudagraphs'} as backends in an effort to improve performance. Dynamo is a Python-level Just-In-Time (JIT) compiler that intercepts Python bytecode execution to extract PyTorch operations into an FX Graph, which can then be optimized and executed by custom backends. However, we observed no significant speedup during training, suggesting that these optimizations may not yet be well-suited for KohyaSS scripts or our specific training.

Despite some optimization frameworks like TorchDynamo not providing significant speedups, KohyaSS scripts offer various optimization techniques to accelerate training and optimize VRAM usage. The training script (Code \ref{code6}) we employed leverages several key optimizations to enhance performance and stability:

\begin{itemize}
    \item \texttt{gradient\_checkpointing}: This technique reduces memory consumption by recomputing activation functions during the backward pass rather than storing them in GPU memory. This allows for training larger models without exceeding VRAM limits, albeit at the cost of slightly increased computation time \cite{bmaltais_lora_training_parameters2023}.
    \item \texttt{highvram}: This optimizes tensors allocation to fully utilize available GPU memory. This is particularly useful for large-scale models, ensuring that memory fragmentation is minimized and computational resources are used efficiently.
    \item \texttt{sdpa} (Scaled Dot-Product Attention): It improves the efficiency of Attention computation on supported hardware by leveraging optimized Key, Query and Values matrices multiplications in Transformers models \cite{huggingface_diffusers_torch2_optimization_sdpa}.
    \item Latent and Text Encoder Output Caching: By enabling \texttt{cache\_latents\_to\_disk} and \texttt{cache\_text\_encoder\_outputs\_to\_disk}, we minimize redundant computations, reducing both training time and memory consumption. This is particularly effective when working with large datasets, as it avoids recomputing intermediate representations for each training iteration \cite{kohya_sd_scripts}.
    \item Selective Network Training: The \texttt{network\_train\_unet\_only} argument ensures that only the U-Net/DiT component of the model is updated during training, while keeping other components such as the text encoders frozen. This significantly reduces the number of trainable parameters, leading to faster convergence and lower VRAM requirements while still allowing effective fine-tuning of the diffusion model \cite{kohya_sd_scripts}.
\end{itemize}

When fine-tuning with LoRA, two key parameters define the network structure and influence on the original model:
\begin{itemize}
    \item Network Rank: Determines the dimensionality of the learned weight updates. A higher rank allows the network to learn more expressive adaptations but requires more VRAM and often more epoch to converge.
    \item Network Alpha: Acts as a scaling factor for these learned weights. In other words, it adjusts how much the LoRA weights contribute to the overall model adjustments.
\end{itemize}

It is crucial that Network Alpha does not exceed Network Rank. While it is technically possible to set an Alpha higher than Rank, doing so often leads to unintended LoRA behavior and instability in training. The Alpha value directly influences the effective learning rate. If Alpha is smaller than Rank, the strength of the LoRA weight updates is reduced \cite{bmaltais_lora_training_parameters2023}. For example: if Alpha = 8 and Rank = 16, then the effective weight strength is $8 \div 16$ = 0.5. This means that the effective learning rate is only half of the original Learning Rate setting.

The other training arguments are kept at their default values as specified in the Flux and KohyaSS repositories, particularly for configurations related to discrete flow shift and learning rate \cite{black_forest_labs_flux1_dev_HF, kohya_sd_scripts}. These defaults ensure compatibility and stability while leveraging the optimizations provided by both frameworks.

\begin{figure}[H]
    \centering
    \includegraphics[width=0.9\linewidth]{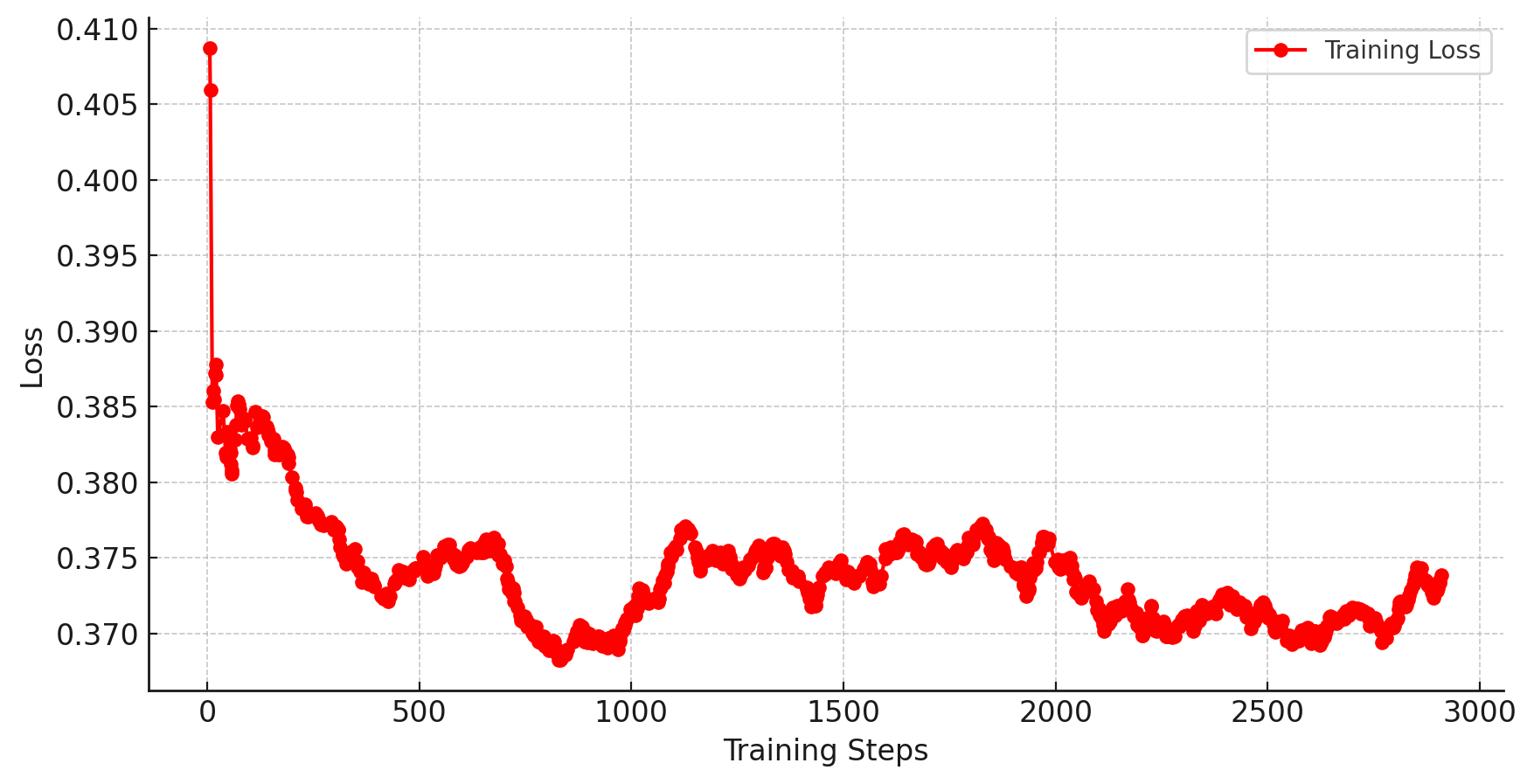}
    \caption{Training loss over steps.}
    \label{fig6}
\end{figure}

With a batch size of 4, a Network Rank (\texttt{network\_dim}) of 8 and a training over 10 epochs, the total training time was approximately 13 hours. As shown in Figure \ref{fig6}, the loss remained relatively stable after an initial sharp decrease, indicating that the model reached a steady state without significant fluctuations. While minor oscillations persisted, they were within an expected range due to batch variability. 

It is also important to note that LoRA training often does not exhibit a strong loss convergence compared to full fine-tuning. This is because LoRA modifies only a subset of parameters (low-rank matrices) rather than the full model, meaning the error may plateau at a higher value while still producing effective results. Despite this, the model’s performance remains robust as long as the loss stabilizes without excessive divergence.

\begin{lstlisting}[float, language=bash, caption={Script and parameters used for LoRA training}, label={code6}]
accelerate launch flux_train_network.py 
    --pretrained_model_name_or_path= #path to FLUX.1-dev 
    --clip_l= #path to Text Encoder 1, CLIP_L
    --t5xxl= #path to Text Encoder 2, T5XXL 
    --ae= #path to Autoencoder
    --dataset_config= #path to dataset .toml file 
    --output_dir= #path to the output directory
    --output_name= #output name
    --save_model_as safetensors 
    --sdpa 
    --highvram 
    --gradient_checkpointing 
    --persistent_data_loader_workers 
    --max_data_loader_n_workers 2 
    --mixed_precision bf16 
    --save_precision bf16 
    --learning_rate 1e-4 
    --max_train_epochs 10 
    --train_batch_size 4 
    --gradient_accumulation_steps 1 
    --network_train_unet_only 
    --network_dim 8 
    --network_alpha 8 
    --network_module networks.lora_flux 
    --cache_latents_to_disk 
    --cache_text_encoder_outputs_to_disk 
    --optimizer_type adamw8bit 
    --timestep_sampling shift 
    --discrete_flow_shift 3.1582 
    --save_every_n_epochs 2 
    --model_prediction_type raw 
    --seed 42 
    --guidance_scale 1.0 
    --log_with tensorboard 
    --logging_dir "/workspace/" 
\end{lstlisting}

\clearpage

\section{Results}\label{sec4}
The proposed reconstruction framework yielded satisfactory results already in the second stage of the process, which leveraged ControlNet for refining image details. However, further improvements were sought by training a custom LoRA to enhance the reconstruction of specific elements such as shirts, logos, and football action scenes.

\begin{figure}[H]
    \centering
    \includegraphics[width=0.95\linewidth]{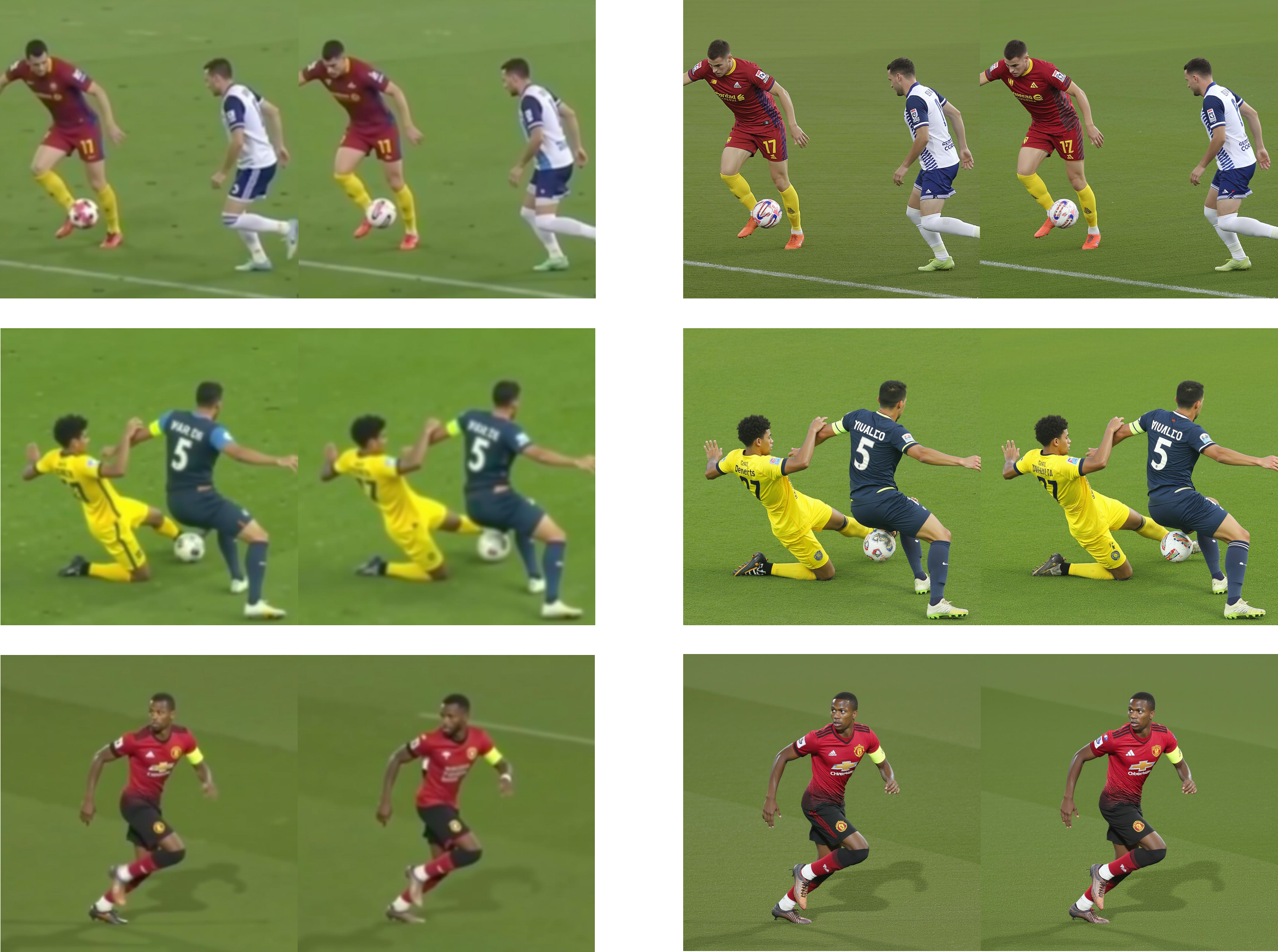}
    \caption{Comparison of reconstruction quality across two stages with and without LoRA fine-tuning. The first column represents the first stage of reconstruction, with the left image being the output without LoRA and the right image incorporating LoRA. The second column represents the second stage of reconstruction (ControlNet), following the same pattern: left without LoRA and right with LoRA.}
    \label{fig7}
\end{figure}

As illustrated in Figure \ref{fig7}, the ControlNet pipeline with the LoRA weights attached demonstrated significant improvements in several aspects of the reconstruction process. Most notably, the clarity and sharpness of player uniforms were enhanced, leading to better visibility of team logos, jersey numbers, and sponsorship placements. Despite these advances, some difficulties remained; in particular, recreating the intricate details of the shirt logo writings proved challenging, even with our trained LoRA. It is important to note that the code for implementing the LoRA remains substantially the same as Code \ref{code3}, except for the addition of the line \texttt{pipe.load\_lora\_weights(cross\_attention\_kwargs={"scale": 0.9})}, which enables developers to control the extent to which the LoRA weights are incorporated into the original model (with a value of 1 being equivalent to using the fully fine-tuned LoRA) \cite{huggingface_diffusers_loading_adapters}.

It is worth noting that the application of our trained LoRA did not uniformly benefit all stages of the reconstruction pipeline. In some cases, especially during the first stage of the reconstruction, the images were better reconstructed without incorporating the LoRA, suggesting that its integration at this early stage could introduce inconsistencies. In contrast, for the second stage of the reconstruction, where ControlNet plays a pivotal role in refining details, applying our trained LoRA consistently provided substantial benefits.

Additionally, finer details, such as hands, shoes, and complex textures, exhibited greater accuracy and definition in the LoRA-enhanced outputs. This improvement was particularly evident in fast-moving actions, where diffusion-based reconstructions traditionally struggle to retain structure due to motion blur. The ability of the LoRA-tuned model to recover these details suggests that the fine-tuning process successfully learned and integrated domain specific visual cues from the football dataset.

\begin{figure}[H]
    \centering
    \includegraphics[width=0.95\linewidth]{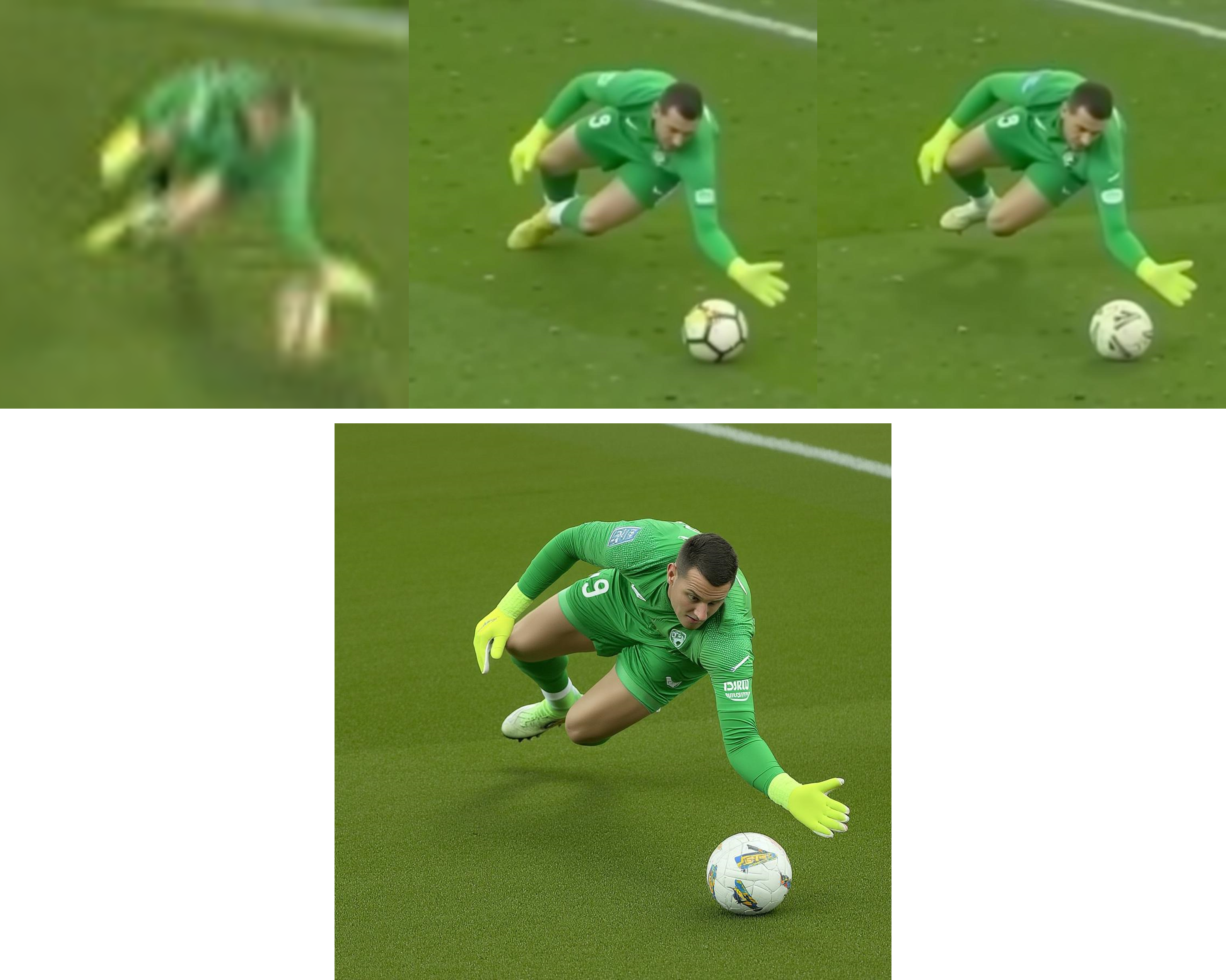}
    \caption{The first row presents the original raw frame at 64×64 resolution, followed by the reconstructed output from the first stage without LoRA, and then the output incorporating LoRA. The second row highlights the final reconstruction result after the second stage (ControlNet), demonstrating the significant enhancement in detail and realism achieved with LoRA and ControlNet.}
    \label{fig8}
\end{figure}

A particularly compelling example of the benefits provided by the trained LoRA can be observed in the reconstruction of a goalkeeper image, Figure \ref{fig8}. As shown, the image generated without using LoRA exhibited noticeable deformities; notably, the two legs of the goalkeeper appeared to fuse into one. In contrast, when employing our trained LoRA, the reconstruction maintained a clear and distinct separation between the two legs. This improvement is especially significant considering the extremely low starting resolution of 64 x 64, which typically poses substantial challenges for preserving fine structural details.

Furthermore, it is worth noting that the LoRAs created by the KohyaSS repository adhere to the Black Forest Lab (BFL) format, which differs from those trained with Diffusers-based repositories \cite{huggingface_diffusers_loading_adapters}. The KohyaSS repository provides a Python script called \textit{convert\_flux\_lora.py} to convert LoRAs between the BFL format and the Diffusers format. We conducted inference experiments with both formats after converting our LoRA into Diffusers, and observed no notable differences in inference speed or output quality, highlighting that in the latest versions of Diffusers ($>0.30.3$), any discrepancies between the formats have largely become negligible.

In summary, the integration of ControlNet and LoRA within our pipeline proved highly effective for football broadcast image reconstruction. The results indicate that leveraging targeted fine-tuning techniques can significantly enhance structural fidelity, particularly for elements that are crucial for accurate and aesthetically convincing sports imagery. It is important to note, however, that despite these advancements, controlling the generated output remains a nuanced challenge. While the modularity of Diffusion Models enhanced by ControlNet and LoRA offers a straightforward pathway to inject domain-specific details, this same flexibility can sometimes lead to unpredictable variations, where minor tweaks in input or parameter settings may result in disproportionate changes in the output. This duality underscores the ongoing need for more robust control strategies to ensure consistent, high-quality reconstructions in complex visual scenarios.

\section{Conclusions}\label{sec5}
This study explored the use of Diffusion Models for reconstructing and enhancing low-resolution football broadcast images. By integrating ControlNet and LoRA fine-tuning, we achieved substantial improvements in image clarity and structural fidelity. Our results show that while ControlNet alone effectively refined general details, the addition of a custom LoRA significantly enhanced the reconstruction of jerseys, logos, and motion elements, ensuring greater precision in uniform textures. These enhancements highlight the effectiveness of domain-specific fine-tuning for specialized image restoration tasks.

Beyond football broadcasts, our findings demonstrate the broader potential of diffusion-based reconstruction techniques in applications requiring high-fidelity image recovery from degraded inputs. In this context, we are considering the development of specialized LoRA modules trained specifically on individual football teams or players. Such targeted fine-tuning is expected to further enhance reconstruction capabilities by capturing team-specific or player-specific visual features, thereby improving the accuracy of restoring details like uniform patterns, emblems, and distinctive motion characteristics.

Moreover, our experiments revealed that in some cases the initial reconstruction stage produced better results without applying LoRA. To address this variability, we are planning to implement a parallel execution strategy that concurrently generates both LoRA-enhanced and non-LoRA outputs. This approach will facilitate rapid comparison and enable the selection of the optimal reconstruction based on the specific characteristics of each input image, enhancing both the efficiency and quality of the process.

In addition, we plan to develop a user-friendly graphical user interface (GUI) to streamline the process and make the pipeline accessible to less technical users. This GUI will simplify the deployment and use of our system, further broadening its practical applicability.

Furthermore, future work will focus on scaling our approach to a larger dataset containing more than 5,000 images, further refining the model’s ability to reconstruct football-related details with higher accuracy. We also plan to explore the possibility of fine-tuning a Vision Model for automatic, offline captioning of dataset images, reducing reliance on external AI services for metadata generation. Finally, we intend to explore the potential for converting these images into three-dimensional model representations using future Img-to-3D models.
By leveraging generative AI with targeted fine-tuning and adaptive reconstruction strategies, this research sets a foundation for next-generation image enhancement techniques, effectively bridging the gap between traditional upscaling methods and AI-driven super-resolution.

\bmhead{Supplementary information}

The supplementary materials include code implementations and output images generated during the reconstruction process. The dataset used in this study can be made available upon specific request.

\section*{Declarations}
\subsection*{Funding}
This work was partially supported by the Ecosystem “RAISE—Robotics and Artificial Intelligence for Socio-Economic Empowerment” (NextGenerationEU code ECS 00000035) PNRR-M4C2-I1.5.
\subsection*{Conflict of interest}
The authors declare no conflict of interests.
\subsection*{Ethics approval}
Ethical approval was not required for this study.
\subsection*{Consent for publication}
Not applicable.
\subsection*{Data availability}
The dataset and image outputs that support the findings of this study are included with the submitted manuscript. They are not publicly available due to proprietary restrictions. However, they can be obtained from the corresponding author upon reasonable request.
\subsection*{Materials availability}
Not applicable.
\subsection*{Code availability}
The dataset and image outputs that support the findings of this study are included with the submitted manuscript. They are not publicly available due to proprietary restrictions. However, they can be obtained from the corresponding author upon reasonable request.
\subsection*{Author contribution}
All authors contributed equally to this work.

\bibliography{sn-bibliography}

\end{document}